\documentclass[10pt,twocolumn,letterpaper]{article}

\usepackage{cvpr}
\usepackage{times}
\usepackage{epsfig}
\usepackage{graphicx}
\usepackage{amsmath}
\usepackage{amssymb}
\usepackage{comment}
\usepackage{color}

\usepackage[breaklinks=true,bookmarks=false]{hyperref}

\cvprfinalcopy % *** Uncomment this line for the final submission

\def\cvprPaperID{****} % *** Enter the CVPR Paper ID here

\title{CBR-Net: Cascade Boundary Refinement Network for Action Detection: \\
Submission to ActivityNet Challenge 2020 (Task 1)}
\author{Xiang Wang$^1$ \quad Baiteng Ma$^1$  \quad Zhiwu Qing$^1$ \quad Yongpeng Sang$^2$ \quad Changxin Gao$^1$ \\
\quad Shiwei Zhang$^{3*}$ \quad Nong Sang$^1$\thanks{
Corresponding authors
}
\\
$^1$School of Artificial Intelligence and Automation, Huazhong University of Science and Technology\\
$^2$School of Cyber Science and Engineering, Huazhong University of Science and Technology\\
$^3$DAMO Academy, Alibaba Group\\
{\tt\small \{u201613707, btm, qzw, ypsang, cgao , nsang\}@hust.edu.cn}\\
{\tt\small zhangjin.zsw@alibaba-inc.com}
}
\begin{document}
\maketitle
%\thispagestyle{empty}
%
%%%%%%%%% ABSTRACT
\begin{abstract}
%-----------------------------------------------------------------------------
%
In this report, we present our solution for the task of \textbf{temporal action localization (detection) (task 1)} in ActivityNet Challenge 2020.
The purpose of this task is to temporally localize intervals where actions of interest occur and predict the action categories in a long untrimmed video. 
Our solution mainly includes three components:
1) feature encoding: we apply three kinds of backbones, including TSN~\cite{TSN}, Slowfast\cite{slowfast} and I3D\cite{I3D}, which are both pretrained on Kinetics dataset\cite{I3D}.
Applying these models, we can extract snippet-level video representations;
2) proposal generation:
we choose BMN~\cite{lin2019bmn} as our baseline, base on which we design a Cascade Boundary Refinement Network (CBR-Net) to conduct proposal detection.
The CBR-Net mainly contains two modules: 
temporal feature encoding, which applies BiLSTM to encode long-term temporal information; 
CBR module, which targets to refine the proposal precision under different parameter settings;
3) action localization:
In this stage, we combine the video-level classification results obtained by the fine tuning networks to predict the category of each proposal.  
Moreover, we also apply to different ensemble strategies to improve the performance of the designed solution, by which we achieve 42.788\% on the testing set of ActivityNet v1.3 dataset in terms of mean Average Precision metrics and achieve Rank 1 in the competition.
\end{abstract}
% 
%%%%%%%%% BODY TEXT
\section{The proposed method}
In this section, we present our solution detailedly.
Firstly, we introduce the applied models for video representation.
Secondly, we present the proposed CBR-Net for proposal generation.
Thirdly, we discuss the ensemble strategies in our solution.

\subsection{Video Representaion}
We first extract features from image frames, and the features will be used as the input of CBR-Net.
%
% Soft-NMS is conducted on the output of CBR-Net to produce the
% final output.

In this section, we extract features from input videos via TSN~\cite{TSN}, I3D~\cite{I3D} and Slowfast~\cite{slowfast} models, which can be formed into a 2D Temporal Feature Sequence. 
Follow the previous works~\cite{bsn, lin2019bmn, DBG, 2D-TAN}, we construct feature sequence under a same temporal scale to make the training operation more efficient.

\textbf{Temporal Segment Network.} Temporal segment network (TSN~\cite{TSN}), a simple but efficient framework for action recognition task, which is based on the idea of long-range temporal structure modeling. 
It combines a sparse temporal sampling strategy and video-level supervision to enable efficient and effective model learning by using the whole action video. 
TSN takes a strategy of sampling a fixed number of sparse segments from one video to model long-term temporal structure. 
In particular, the final prediction of video is averaged by the logits of each chip. 
In this competition, we experimented with temporal segment network by sampling 16 frames for each clip.
\begin{figure*}
\begin{center}
\end{center}
   \includegraphics{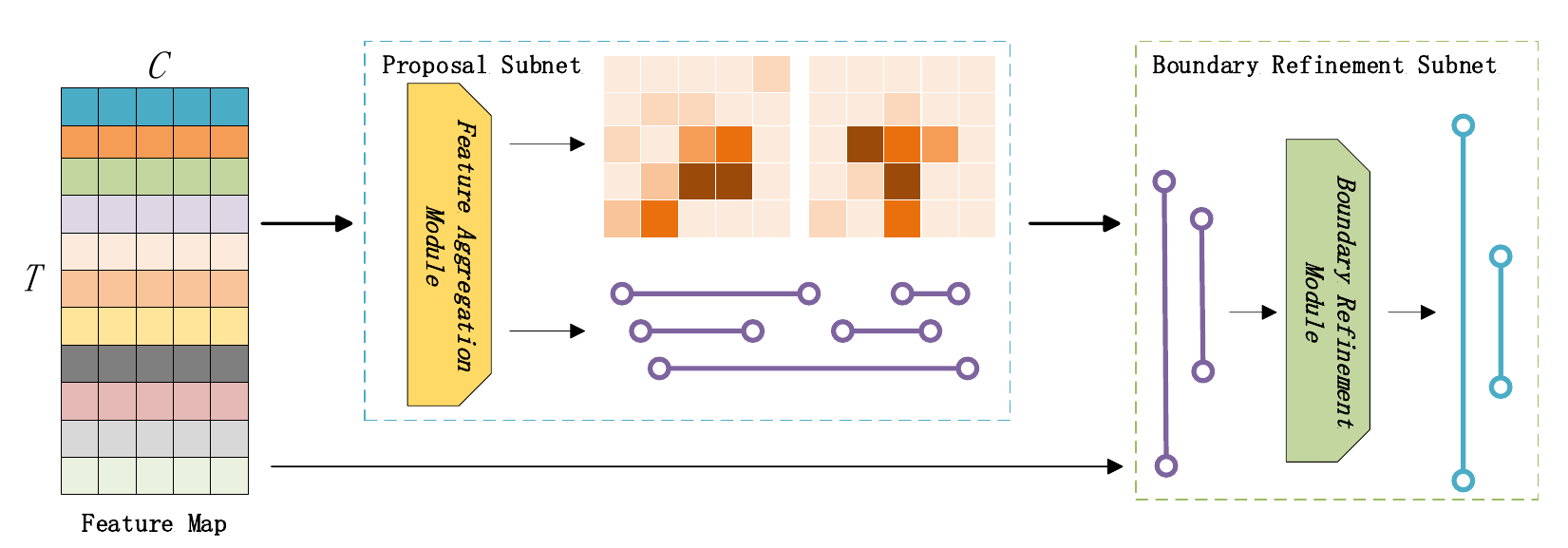}
   \caption{ The framework of CBR-Net. 
   First, the extracted features are the input of the Proposal Subnet.
   In this stage, we target to generate proposals with high recall rate. 
   Then the obtained proposals and the extracted features are input into Boundary Refinement Subnet together for fine-tuning the boundary.}
\label{fig:netwok}
\end{figure*}

\textbf{Inflated 3D ConvNet.}
Inflated 3D ConvNet (I3D~\cite{I3D}) is based on 2D ConvNet inflation: 
filters and pooling kernels of 2D CNN are expanded to 3D, making it possible to learn seamless Spatio-temporal feature extractors from video while leveraging successful ImageNet pretrained architecture designs and even their parameters.
In this competition, we experimented with the Inflated 3D ConvNet with sampling 16 frames form one clip to extract the features. 
At the same time, we use Kinetics pretrained model to initialize our I3D network.

\textbf{Slowfast Network.}
Slowfast network (Slowfast~\cite{slowfast}), which involves (i) a Slow pathway, operating at a low frame rate, to capture spatial semantics, and (ii) a Fast pathway, operating at a high frame rate, to capture motion at fine temporal resolution. 
The Fast pathway can be made very lightweight by reducing its channel capacity, yet it can learn useful temporal information for video recognition. For details about the architecture, please refer to its raw publication~\cite{slowfast}.
In particular, in this competition, for the input videos, we keep the frame rate at 15FPS to extract video frames. At the same time, each clip contains 32 frames is input to slowFast network~\cite{slowfast}.

\subsection{Cascade Boundary Refinement Network}

% 总写，本段介绍CBR-Net，主要包括Temporal Encoding 和 Cascade Module。。 以BMn做baseline，本段先介绍 BMN，然后介绍时序编码和边界修正。
% 
In this section, we introduce the CBR-Net designed in the competition, as is shown in Figure ~\ref{fig:netwok}. 
CBR-Net mainly contains two components: temporal feature encoding module and CBR module, which will be introduced in this section detailly.
Actually, the CBR-Net is designed based on the Boundary Matching Network (BMN~\cite{lin2019bmn}), thus we will present the overview of BMN first to make the report easier to be understood.
%
% We will mainly introduce Temporal Feature Encoding module and Cascade Boundary Refinement (CBR) module in our CBR-Net. At the same time, Boundary Matching Network (BMN\green{~\cite{lin2019bmn}}) will be introduced as the baseline. 
% 
% For the sake of description and explanation, we introduce BMN first, then Temporal Feature Encoding module and Cascade Boundary Refinement module.
%1  BMN

\textbf{Boundary Matching Network.}
BMN is mainly composed of two modules: temporal evaluation module and proposal evaluation module. 
The goal of the temporal evaluation module is to evaluate the starting and ending probabilities for all temporal locations in the untrimmed video by constructing two temporal 1D convolutional layers on feature maps. 
These boundary probability sequences are used for generating proposals during post-processing. 
The goal of the proposal evaluation module is to generate Boundary-Matching (BM) confidence map, which contains confidence scores for densely distributed proposals. 
Here, the BM confidence map takes the starting time points of action as its x-coordinate and the during of action as its y-coordinate. 
The temporal evaluation module and proposal evaluation module are jointly trained in a unified framework in BMN.
%
%2 LSTm

\textbf{Temporal Feature Encoding module.}
Temporal Feature Encoding module is a simple but effective module in Our CBR-Net. The goal of this module is to encoding temporal information, which mainly by constructing BiLSTM layers on the feature map to excavate the relationship between different time points in the feature map.
% 
% 3 Casecade

\begin{figure}[t]
\begin{center}
%   \includegraphics[width=0.8\linewidth]{egfigure.eps}
\includegraphics[width=8cm]{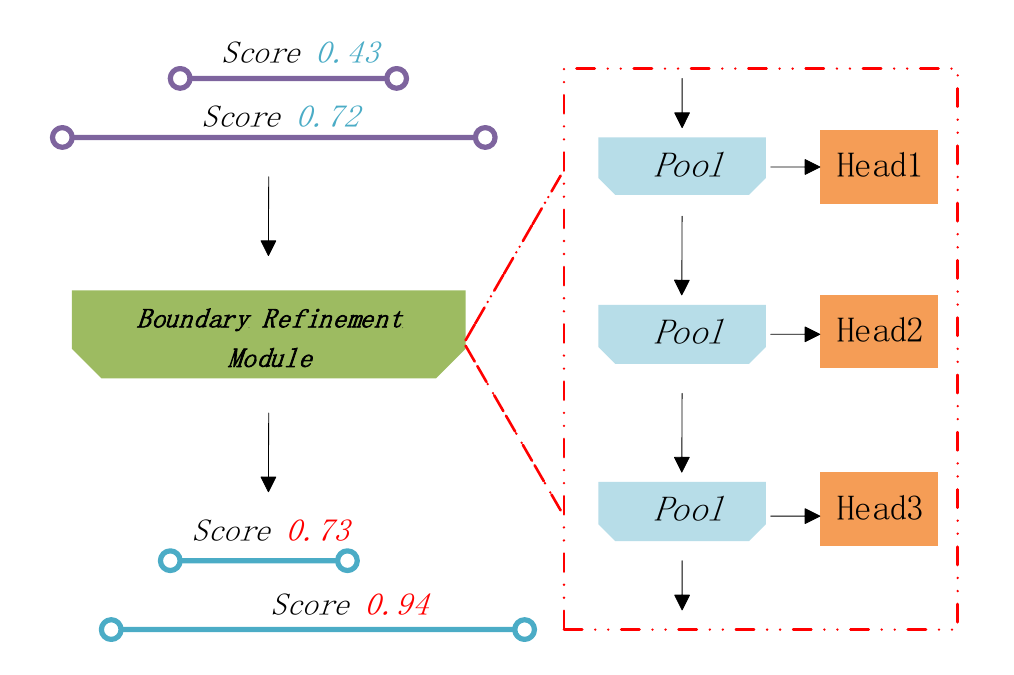}
\end{center}
  \caption{Details of Boundary Refinement Subnet. For the input proposals, this subnet outputs the results of fine-tuning the boundary and confidence through the multi-layer cascade structure. Here, a three-layers structure is shown.}
\label{fig:refinement}
\end{figure}

\textbf{Cascade Boundary Refinement module.}
By analyzing the BMN network, we can find that the length and boundary of the proposals output by BMN are fixed. And to obtain a more precise boundary, the size of the BM confidence map in BMN needs to be increased, which poses a challenge to the computation and makes BMN inefficient. 
To solve this problem, we proposed Cascade Boundary Refinement (CBR) module. CBR module takes the coarse-boundary proposals output from Proposal Subnet as input, and the details of the CBR module is shown in figure~\ref{fig:refinement}. Obviously, the goal of the CBR module is to output proposals with finer boundaries.
% 
% 
\begin{comment}

In the competition, we used a new network proposed by us.
%
The schematic diagram of this network is shown in Figure~\ref{fig:netwok}.\\
\indent Temporal-Aware Reﬁnement Network (TAR-Net) consists of two subnets : Proposal Subnet and Boundary Refinement Subnet. Proposal Subnet aims to obtain a proposal with a high recall rate and a reliable confidence score, while Boundary Refinement Subnet aims to fine-tune the boundary to make the proposal boundary more accurate. The Boundary Refinement Subnet is shown in the Figure~\ref{fig:refinement}.\\
% 
\indent At the same time, the experimental results show that our network is very effective, and our competition is mainly based on this network to get the final detection results.
\end{comment}

\subsection{Ensemble Strategies}
In the competition, we use our Cascade Boundary Refinement Network (CBR-Net) and the previous state-of-the-art works, Boundary Sensitive Network (BSN~\cite{bsn}) and Boundary Matching Network (BMN~\cite{lin2019bmn}) to conduct model ensemble. 
Then we integrate all of model results to get the final results. At the same time, the multi-feature fusion strategy is also used in the competition.
In particular, we found that the ensemble strategies are greatly effective for improving detection performance in the competition.
\begin{table}
\begin{center}
\begin{tabular}{|l|c|c|}
\hline
Method & Validation(mAP) &  Testing(mAP) \\
\hline\hline
BSN(baseline) & 30.03 & 32.84 \\
BSN & 32.8 & -\\
BMN(baseline) & 33.85 & 36.42\\
BMN & 36.5 & -\\
\bf CBR-Net & \bf 38.0 & - \\
\hline\hline
\bf Ensemble & \bf 40.1 & \bf 42.788\\
\hline
\end{tabular}
\end{center}
\caption{Temporal action detection results. Performance comparison between models and the final test results.}
\label{tab:results}
\end{table}
% 
%-------------------------------------------------------------------------
% 
\section{Experiments}
\subsection{Dataset}
ActivityNet-1.3 is a large dataset for general temporal action detection, which contains 19994 videos with 200 action classes annotated and was used in the ActivityNet Challenge 2016, 2017, 2018, and 2019. 
ActivityNet-1.3 is divided into training, validation, and testing sets by ratio of 2:1:1. 
In the competition, we train the model on the original partitioned training set and verify the model results on the original partitioned validation set. 
The result of the competition is the result of the model on the testing set. The label for the testing set is not disclosed.
% \url{http://www.pamitc.org/documents/mermin.pdf}.
% 
\subsection{Metrics}
In the temporal action detection task, mean Average Precision (mAP) is adopted as the evaluation metric, where we calculate
Average Precision (AP) on each action category respectively. Mean mAP with IoU thresholds $[0.5 : 0.05 : 0.95]$ are used on ActivityNet-1.3.
\subsection{Experiment Results}
\begin{figure} %[t]
\begin{center}
%   \includegraphics[width=0.8\linewidth]{egfigure.eps}
\includegraphics[width=8cm]{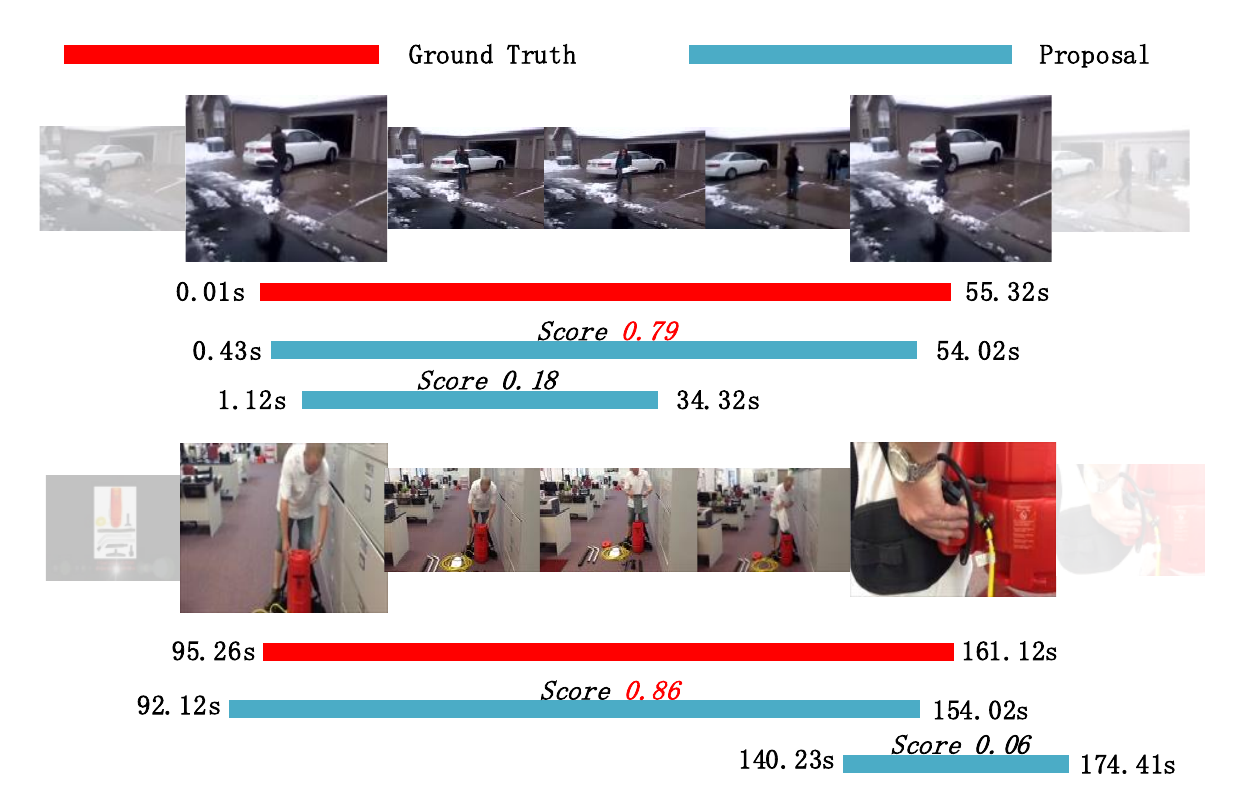}
\end{center}
  \caption{Visualization examples of proposals on ActivityNet-1.3 dataset.}
\label{fig:quality}
\end{figure}

In this section, we compared the performance of several current advanced models with that of our CBR-Net in the competition. As Shown in Table~\ref{tab:results}, it should be noted that the baseline in parenthesis means the result in the original paper, the baseline can be greatly improved by using our feature extraction and some model training and fusion tricks. Among them, BSN improved by 2.77\% on mAP compared to the original paper, and BMN improved by 2.65\% on mAP compared to the original paper.
In particular, we argue that our proposed CBR-Net is well suited for the task of temporal action detection, and it turns out that our CBR-Net exceeds the current state of the art models, achieved 38.0\% mAP of performance. Figure~\ref{fig:quality} shows examples of proposal generated by our CBR-Net.\\
\indent In the competition, on the testing set, our final results integrated the results of BSN, BMN, and CBR-Net. And results finally reached 42.788\% mAP.
\subsection{Analysis}
From the results in Table~\ref{tab:results}, we can find that CBR-Net can output state-of-the-art method by 1.5\%, which can better demonstrate the effectiveness of the proposed method.
Moreover, in the ensemble stage, we can apply the complementarity among different methods, \emph{i.e.} BMN, BSN and CBR-Net, to improve the detection performance.
%
% This work is built upon the ideas of fusion features of diversity and using CBR-Net to produce high-quality proposals.
% 
% \indent In the competition, we 

% 
%------------------------------------------------------------------------
%-------------------------------------------------------------------------
% \subsection{Margins and page numbering}
% 
% 
% 
%-------------------------------------------------------------------------
\section{Conclusion}
In this work, we propose a novel action detection network enhanced with temporal encoding module and CBR module for temporal action detection task. 
Especially, the experimental results show that the solution CBR-Net can significantly improves the detection performance. 
At the same time, in the competition, we also ensemble some other previous networks for better performance.
%-------------------------------------------------------------------------
% % 
% \begin{table}
% \begin{center}
% \begin{tabular}{|l|c|c|}
% \hline
% Method & Validation(mAP) &  Testing(mAP) \\
% \hline\hline
% BSN(baseline) & 30.03 & 32.84 \\
% BSN & 32.8 & -\\
% BMN(baseline) & 33.85 & 36.42\\
% BMN & 36.5 & -\\
% \bf TAR-Net & \bf 38.0 & - \\
% \hline\hline
% \bf Ensemble & \bf 40.1 & \bf 42.788\\
% \hline
% \end{tabular}
% \end{center}
% \caption{Temporal action detection results. Performance comparison between models and the final test results.}
% \label{tab:results}
% \end{table}
% % 
%-------------------------------------------------------------------------

{\small
\bibliographystyle{ieee}
\bibliography{egbib}
}

\end{document}